\documentclass{article}

\usepackage{arxiv}

\usepackage[utf8]{inputenc}
\usepackage[T1]{fontenc}
\usepackage{amsmath}
\usepackage{amssymb}
\usepackage{amsfonts}
\usepackage{graphicx}
\usepackage{hyperref}
\usepackage{url}
\usepackage{booktabs}
\usepackage{nicefrac}
\usepackage{microtype}
\usepackage{natbib}
\usepackage{doi}
\usepackage{xcolor}
\graphicspath{{figures/}{./}}

\title{Eviction as Estimation: A Fixed-Lag Smoothing View of Test-Time Memory, and When Measuring Beats Accumulating}
\date{}

\author{%
  \href{https://orcid.org/0009-0003-3503-7321}{\includegraphics[scale=0.06]{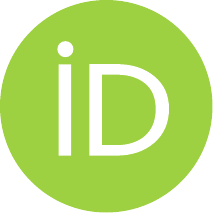}\hspace{1mm}Maruthi Vemula} \\
  University of North Carolina at Chapel Hill \\
  \texttt{vemula@unc.edu} \\
  \And
  \href{https://orcid.org/0009-0007-7967-6941}{\includegraphics[scale=0.06]{orcid.pdf}\hspace{1mm}Neeraj Praneeth Gajula} \\
  New York University \\
  \texttt{Npg8461@nyu.edu} \\
}

\renewcommand{\shorttitle}{Eviction as Estimation}

\hypersetup{
  pdftitle={Eviction as Estimation: A Fixed-Lag Smoothing View of Test-Time Memory},
  pdfauthor={Maruthi Vemula, Neeraj Praneeth Gajula},
  pdfkeywords={KV cache compression, test-time memory, attention, long-context inference, cache eviction},
}

\begin{document}

\maketitle

\begin{abstract}
A language model with a bounded working memory, whether a key-value cache or a recurrent state, must repeatedly decide which stored items to keep and which to discard. Every deployed method makes this choice at the moment an item arrives, either from the past (recency in StreamingLLM, accumulated attention in H2O) or from a guess about the future (learned predictors such as SnapKV). We recast the choice as an estimation problem on a hidden signal, namely whether an item will be reused, and this view places existing methods on a single axis, the commit lag $H$: online filters and learned predictors both commit at $H=0$, while the classical offline optimum of Belady sits at the opposite end where the whole future is known. The regime in between, which we call fixed-lag smoothing, has not been used for neural memory. Its rule is to wait a bounded number of steps, observe which items a correct near-future prediction actually attended to, and only then commit. This measurement, which we call demonstrated utility, turns Belady's unobservable future request into something we can read off the model itself. We instantiate the idea as a training-free policy, RMM, that is a strict generalization of H2O and reduces to it exactly when the measurement is uniform. We report both sides of the evidence. In controlled settings where reuse is endogenous and clearly separated in time, demonstrated utility identifies used memory far better than accumulated attention, and a small bounded memory behaves like a much larger one. On independent third-party benchmarks, however, the advantage mostly disappears: run inside NVIDIA's KVPress harness against its own SnapKV, H2O, and StreamingLLM implementations, RMM is on par with H2O for single-turn question answering and loses to both H2O and SnapKV in a streaming multi-turn setting. We trace this to a simple cause: on natural text the model is correct about most tokens, so weighting attention by correctness barely changes it, and demonstrated utility collapses onto accumulated attention except when reuse is sharp and endogenous, a regime standard benchmarks do not contain. Our contribution is a framework, a training-free policy that is provably H2O-class in general, and an honest map of when measuring beats accumulating, rather than a new state of the art.
\end{abstract}


\section{Introduction}

A model reading a long input under a fixed memory budget faces the same question at every step: keep this item, or evict it? Keep too little and it loses information a later query needs. Keep everything and it runs out of budget. This trade-off is the core of key-value cache compression, streaming attention, and long-horizon agent memory, and it is only becoming more pressing as context windows and agent sessions grow.

Every method we are aware of answers the question the moment the item arrives. StreamingLLM keeps the most recent tokens \citep{xiao2024streaming}. H2O keeps the tokens that have accumulated the most attention so far \citep{zhang2023h2o}. A more recent line of work trains a model to predict which items will be needed and evicts on that prediction \citep{li2024snapkv,cai2024pyramidkv}. Some methods score from the past and some from a guessed future, but they share one commitment: the decision is made immediately.

The difficulty is that utility is a property of the future. Whether an item matters is revealed only when something downstream uses it, when a fact is referenced again, an intermediate result is reused, or a tool output is read a second time. At the moment an item arrives that evidence does not exist yet, so an immediate decision is a guess, and the recent learned-prediction work is an attempt to guess better.

We take a different route, borrowed from estimation theory. Estimating a hidden state from a stream of observations admits three stances. Filtering estimates the state now, from data up to now. Prediction estimates a future state. Smoothing estimates a past state using data up to a later time, which means it waits and then looks back with more evidence \citep{kalman1960,rauch1965}. The eviction literature does filtering and prediction. Smoothing is missing. Our thesis is that bounded-memory eviction should be done by fixed-lag smoothing: delay each keep-or-evict decision by a bounded lag $H$, and by then the near-future usage has been observed rather than predicted.

This single move has three consequences. First, it unifies the field. The commit lag $H$ is a new axis on which online filters and learned predictors are the $H=0$ special case, and the offline optimum of \citet{belady1966} is the $H\to\infty$ limit. Second, it connects neural memory to provably optimal caching. Belady's algorithm is optimal but needs the future request sequence, which does not exist explicitly in a neural memory. We construct one. Third, and this is the part we want to be careful about, the idea has a regime. When reuse is endogenous and separated in time, measuring near-future utility beats guessing it. When it is not, which is the common case for single-shot long-context reading, the measurement collapses onto accumulated attention and our policy becomes H2O. We establish this boundary with both controlled experiments and independent benchmarks, and we do not hide the negative side.

The bridge to Belady is the technical heart of the paper. Belady needs to know which items the future requests. In a neural memory there is no explicit request stream, so we define one. We call it demonstrated utility: a label-free measurement, read from the model's own attention and next-token correctness, of which cached items a correct near-future prediction actually used. A fixed lag turns this from an unobservable prediction problem into an observable measurement problem.

We state the outcome plainly at the start, because the honest result is the point of the paper. The framework is a genuine contribution. The policy it produces, which we call RMM, is a strict generalization of H2O. On the controlled tasks we built to isolate the mechanism, RMM tracks an offline oracle while filters collapse. On independent benchmarks that we did not build, run inside a third-party harness against that harness's own baselines, RMM is on par with H2O and does not beat it, and in the streaming setting it was designed for it in fact loses to H2O and SnapKV. We explain why, and we argue that the useful contribution is the framework together with a clear account of when the idea helps.

\paragraph{Contributions.}
(1) A framework that recasts bounded-memory eviction as estimation, with the commit lag $H$ as the axis that unifies filters, predictors, and Belady (Section~\ref{sec:framework}).
(2) The Belady bridge: demonstrated utility, a label-free signal that makes Belady's future request observable inside a neural memory (Section~\ref{sec:du}).
(3) RMM, a training-free policy that is a strict generalization of H2O (Section~\ref{sec:rmm}), with a small model that predicts the ordering we then measure (Section~\ref{sec:theory}).
(4) Honest evidence of both signs: controlled settings where the mechanism works (Section~\ref{sec:parta}) and independent benchmarks where it is only H2O-class (Section~\ref{sec:partb}), with an analysis of the cause and a resulting map of when measuring beats accumulating (Section~\ref{sec:discussion}).

\section{Related work}

We organize prior work by what information the eviction decision uses, which is the axis our framework introduces.

\paragraph{Online filters that decide from the past.}
StreamingLLM keeps attention sinks and a recent window \citep{xiao2024streaming}. H2O keeps heavy hitters, the tokens with the most accumulated attention \citep{zhang2023h2o}. Scissorhands \citep{liu2023scissorhands} and FastGen \citep{ge2024fastgen} refine which tokens to keep from past-attention statistics. Surprise-based memories such as Titans keep the most surprising writes \citep{behrouz2024titans}. In our terms these are all $H=0$ policies that differ only in the scoring function.

\paragraph{Learned predictors that decide from a guessed future.}
SnapKV selects key-value pairs using the attention of a recent observation window as a proxy for what a query will need \citep{li2024snapkv}, and PyramidKV allocates budget across layers \citep{cai2024pyramidkv}. These improve the estimator but not the stance: they still commit at arrival, before the evidence of use exists.

\paragraph{The offline optimum.}
Belady's algorithm is the optimal cache replacement policy, but it requires the future request sequence \citep{belady1966}. It is standard in systems caching, where requests are explicit addresses, and it has been used as an unreachable skyline. Its assumption of a known future has no direct neural analogue, and supplying one is our bridge.

\paragraph{Estimation theory.}
Filtering, prediction, and smoothing are the three classical stances for estimating a hidden state from a stream \citep{kalman1960}, and fixed-lag smoothing, which waits a bounded number of steps before committing an estimate of a past state, is the one the eviction literature has not used \citep{rauch1965}. We import it as the organizing idea.

\paragraph{Memory in learning systems.}
Complementary learning systems theory argues that the brain does not decide at encoding time what to keep forever, but consolidates after a delay based on what proved relevant \citep{mcclelland1995cls,kumaran2016cls}. We use this as motivation rather than evidence. Our commit lag is a computational analogue of delayed consolidation, and the key-value cache is simply its most measurable instance.

\paragraph{Benchmarks and harnesses.}
We evaluate on LongBench \citep{bai2024longbench} and LoCoMo \citep{maharana2024locomo}, and we discuss RULER \citep{hsieh2024ruler} and SCBench \citep{li2025scbench}, the last of which is built specifically for key-value methods in multi-turn settings. For the independent comparison we run inside NVIDIA's KVPress harness \citep{kvpress2024}, which ships reference implementations of SnapKV, H2O, and StreamingLLM, so that the baselines are not ours. All experiments use frozen Qwen2.5 models \citep{qwen2025}.

\section{Eviction as estimation}
\label{sec:framework}

\paragraph{Setup.}
Items $x_1, x_2, \dots$ arrive one per step. A bounded memory holds at most $K$ committed items. At each step a policy decides what to keep. Item $x$ has a latent utility $u(x)$, informally how much keeping it reduces future loss on downstream queries. The keep-or-evict decision is an estimate $\hat u(x)$, and methods differ in what $\hat u$ conditions on. Table~\ref{tab:frame} lays out the four classes.

\begin{table}[t]
\caption{Existing eviction methods and ours, organized by the information the decision uses. The commit lag $H$ is the axis: filters and predictors are $H=0$, Belady is $H\to\infty$.}
\label{tab:frame}
\centering
\small
\begin{tabular}{@{}llll@{}}
\toprule
class & conditions on & stance & members \\
\midrule
online filter & past and present ($\le t$) & filter & StreamingLLM, H2O, Titans \\
learned predictor & a learned guess of the future & predictor & SnapKV, PyramidKV \\
fixed-lag smoother (ours) & measured near-future over $[t, t{+}H]$ & smoother & RMM \\
offline optimal & all future & Belady & oracle skyline \\
\bottomrule
\end{tabular}
\end{table}

\begin{figure}[!htb]
\centering
\includegraphics[width=0.80\linewidth]{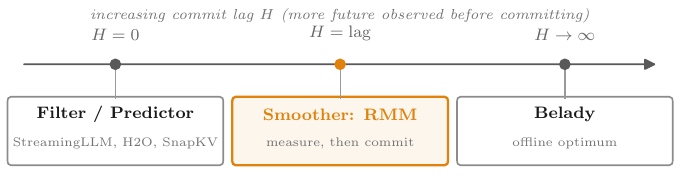}
\caption{The commit-lag axis. Online filters and learned predictors both decide at $H=0$, from the past or a guessed future. Belady decides with the whole future known. RMM sits in between: it waits a bounded lag, measures which items a correct prediction used, and then commits.}
\label{fig:axis}
\end{figure}

\paragraph{The commit lag is a design axis.}
Prior methods all commit at arrival, so they are $H=0$. Sweeping $H$ from $0$ to $\infty$ moves continuously from the online filter to Belady, as sketched in Figure~\ref{fig:axis}. The claim is not that larger $H$ is always better in practice, since the lag costs buffer memory and latency. The claim is that $H$ is the right axis, and that the interesting middle of it has been skipped.

\section{Demonstrated utility: a bridge to Belady}
\label{sec:du}

Belady needs to know which items the future requests. We define an observable surrogate. For a cached item $k$, its demonstrated utility over a lag-$H$ window is the attention it receives from correct near-future predictions,
\begin{equation}
\mathrm{DU}_H(k) \;=\; \sum_{t' \,\in\, (t_k,\, t_k + H]}
   \mathbb{1}\!\left[\text{prediction at } t' \text{ correct}\right]\; a(t' \!\to\! k),
\end{equation}
where $a(t'\!\to\!k)$ is the attention from position $t'$ to item $k$, and correctness is checked by teacher forcing, meaning the model's top prediction at $t'$ equals the realized next token. This is self-supervised and uses no labels. The reading is simple: an item has demonstrated utility if a correct near-future prediction actually used it. A fixed lag $H$ is what makes this quantity measurable, since we only need to wait until the window has passed.

The contrast with H2O is worth stating exactly, because it is the whole story of the paper. H2O scores an item by the total attention it has received, $\sum_{t'} a(t'\!\to\!k)$. Demonstrated utility is the same sum with each term weighted by whether the prediction at $t'$ was correct. So the two are identical except that demonstrated utility ignores attention coming from positions where the model was wrong. When every prediction is treated as correct, demonstrated utility equals H2O. The difference is exactly the correctness weighting, and Section~\ref{sec:partb} shows that on natural text this difference is small.

\section{RMM: a fixed-lag smoothing policy}
\label{sec:rmm}

Each arriving item enters a provisional buffer of size $H$ and is not yet committed. While in the buffer it can still be read, so requests that fall within the lag are served. At age $H$ the item is committed if its demonstrated utility outranks the weakest committed item, and evicted otherwise. The committed cache holds $K$ items and the buffer holds $H$, for a total of $K+H$, an overhead we account for honestly.

At inference on a frozen model, demonstrated utility is computed directly from attention and correctness, so RMM is training-free and drops onto a key-value cache in the same slot occupied by H2O, SnapKV, and StreamingLLM. Because demonstrated utility is H2O's score with a correctness weight, RMM is a strict generalization of H2O: with a uniform weight it reproduces H2O exactly. This is a useful safety property, since it means RMM is never worse than H2O by construction when the correctness signal carries no information, and we verify the exact reduction as a control in Section~\ref{sec:partb}.

\section{A small model of delayed utility}
\label{sec:theory}

We give a minimal model in which the smoother is optimal and every online policy is bounded away from it, and which predicts the ordering we later measure. We present it as an illustration of the mechanism, not as a claim about real text.

\paragraph{Model.}
$N$ items arrive one per step. A committed cache holds $K$ and a provisional buffer holds the most recent $H$. Of the items, $U$ are useful and the rest are useless. Each useful item $i$ has a signal request at some delay $d_i \le H$, which is the event of it being used, and a far test request at time $T \gg H$. Useless items are never requested. Crucially, at arrival a useful item and a useless item look identical, and the only evidence is the signal, which arrives after the item. The objective is test recall, the fraction of useful items still in the committed cache at time $T$.

\paragraph{Two facts.}
A lag-$H$ smoother that commits item $i$ exactly when it has signalled within its buffer window commits all and only the useful items, and so reaches recall $\min(1, K/U)$, which is the Belady optimum under capacity $K$. Any online policy that must commit at arrival has no information separating useful from useless items, so it retains useful items only in proportion $K/N$, and its recall goes to zero relative to the optimum as the number of useless items grows. A learned predictor of accuracy $p$ interpolates between these two, reaching the smoother only as $p \to 1$, which cannot happen when utility is not determined by arrival-time features, and that is the premise. With $K/N = 0.33$ the online floor is about $0.33$, and the measured filters in our controlled simulator sit near this floor while the smoother and the oracle sit near $1.0$, which is the ordering the model predicts.

\section{Controlled experiments}
\label{sec:parta}

We separate the evidence into two parts and ask the reader to weight them accordingly. This section reports controlled settings that we built to isolate the mechanism. They show what the idea does when its assumptions hold, and in several of them we constructed the task so that the effect would appear. They are mechanism analysis, not competitive claims. Section~\ref{sec:partb} is the load-bearing comparison.

\paragraph{Reasoning that drives its own usage.}
On a frozen Qwen2.5-7B model, we ask a multi-hop question whose answer requires chaining two facts among fourteen distractors, and we measure demonstrated utility from the model's own chain of thought, with nothing inserted to mark which facts matter. When the model reasons correctly, demonstrated utility places both gold facts in the top few of sixteen about $0.8$ of the time, against about $0.1$ for accumulated attention and recency. The signal survives natural usage in this probe. Whether it turns into a benchmark win is a separate question, answered in Section~\ref{sec:partb}.

\begin{figure}[!htb]
\centering
\includegraphics[width=0.46\linewidth]{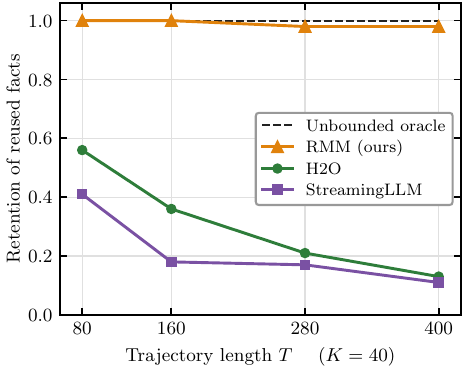}
\caption{A controlled streaming agent. Retention of reused facts against trajectory length at a fixed memory budget $K=40$. RMM holds near the unbounded oracle, while H2O and StreamingLLM decay as the trajectory outgrows the cache. This is a constructed task in which the reused facts carry a clear signal, included to show the mechanism, not as a benchmark result.}
\label{fig:controlled}
\end{figure}

\paragraph{A bounded memory that runs past its budget.}
We stream facts past a memory bounded to $K=40$ items, where a subset are reused shortly after they appear and the rest never recur, and we ask which policy still retains the reused facts after a long trajectory. Figure~\ref{fig:controlled} shows the result. RMM stays near full retention across a trajectory ten times its budget, while accumulated attention and recency decay toward the $K/T$ floor as old facts fall out of the window. In this construction a bounded memory under RMM answers a re-query at about $0.70$ against an unbounded ceiling of $0.80$, while recency answers $0.00$. The effect is real in this setting, where the reused facts carry an explicit reminder. Section~\ref{sec:partb} shows it does not survive to independent benchmarks, where reuse is not so cleanly separable.

\section{Independent benchmarks}
\label{sec:partb}

Everything above is ours, including tasks we tuned to show the effect. The question that decides the paper is whether RMM beats the deployed baselines on code and data we did not write. We implement RMM as a scorer inside NVIDIA's KVPress harness \citep{kvpress2024} and compare against KVPress's own implementations of SnapKV, H2O, and StreamingLLM. As noted in Section~\ref{sec:rmm}, our scorer is KVPress's H2O scorer with one change, a correctness weight on each attention row, so with a uniform weight it reproduces H2O's output exactly. We verified this equality as a control, which means any difference between RMM and H2O below is due only to the correctness weighting. All runs use a frozen Qwen2.5-7B with eager attention at a matched token budget.

\paragraph{Single-turn prefill compression.}
Table~\ref{tab:prefill} reports LongBench \citep{bai2024longbench} question answering and classification at two compression levels. RMM is statistically indistinguishable from H2O: slightly behind on HotpotQA, slightly ahead on TREC at the more aggressive budget, which is its one clear advantage. It does reliably beat SnapKV and StreamingLLM at the aggressive budget. This is what the framework predicts for a single-turn setting, where the compression happens before any answer exists, so there is no correct-use event for demonstrated utility to read and the correctness weight is a small perturbation on H2O.

\begin{table}[t]
\caption{LongBench single-turn prefill compression, frozen Qwen2.5-7B, $n=40$ examples per cell. RMM is on par with H2O and beats SnapKV and StreamingLLM at the aggressive budget. It does not improve on H2O overall.}
\label{tab:prefill}
\centering
\small
\begin{tabular}{@{}llcccc@{}}
\toprule
task & budget & SnapKV & StreamingLLM & H2O & RMM (ours) \\
\midrule
HotpotQA (F1) & keep 50\% & 0.287 & 0.263 & \textbf{0.309} & 0.299 \\
HotpotQA (F1) & keep 25\% & 0.197 & 0.314 & \textbf{0.316} & 0.293 \\
TREC (acc.)   & keep 50\% & 0.600 & 0.625 & 0.725 & 0.725 \\
TREC (acc.)   & keep 25\% & 0.550 & 0.425 & 0.550 & \textbf{0.600} \\
\bottomrule
\end{tabular}
\end{table}

\paragraph{Streaming multi-turn, the setting RMM was designed for.}
Here the history grows turn by turn. Before each turn we re-compress the whole history to a fixed budget using each method's own score, then answer, with the same protocol for every method. The first turn has no prior answer to read, so it is out of regime and RMM equals H2O, which we confirm. Later turns are where a smoother should pay off. Table~\ref{tab:locomo} reports the later-turn result on LoCoMo \citep{maharana2024locomo}. RMM loses, to both H2O and SnapKV. Keying the cache on what earlier correct answers attended to biases it toward parts of the conversation that the next, independent question does not need.

\begin{table}[t]
\caption{LoCoMo streaming multi-turn, frozen Qwen2.5-7B, budget 1800 tokens, $n=8$ conversations, in-regime (later) turns only. RMM loses to both H2O and SnapKV in the very setting it was designed for.}
\label{tab:locomo}
\centering
\small
\begin{tabular}{@{}lccccc@{}}
\toprule
LoCoMo, later-turn F1 & full cache & SnapKV & H2O & RMM (ours) & StreamingLLM \\
\midrule
score & 0.218 & \textbf{0.209} & 0.176 & 0.147 & 0.078 \\
\bottomrule
\end{tabular}
\end{table}

\begin{figure}[!htb]
\centering
\includegraphics[width=0.70\linewidth]{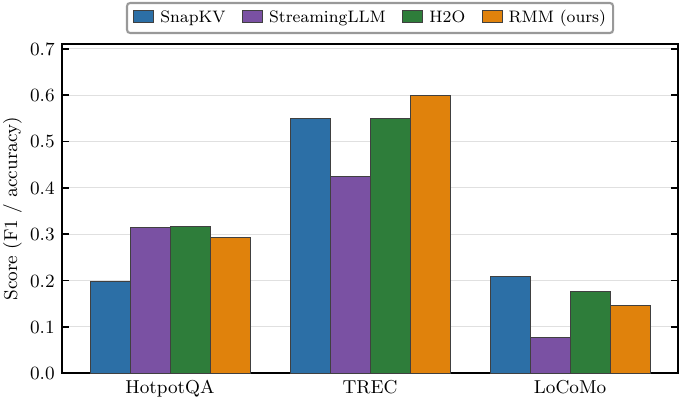}
\caption{Independent-benchmark scorecard inside the KVPress harness (HotpotQA and TREC are single-turn prefill at 25\% budget; LoCoMo is streaming multi-turn, later-turn F1). RMM matches H2O on prefill and loses to both H2O and SnapKV in the streaming setting. The correctness weighting that distinguishes RMM from H2O does not help on these benchmarks.}
\label{fig:independent}
\end{figure}

Figure~\ref{fig:independent} collects these results. Against H2O, the strongest baseline, RMM wins one condition, ties one, and loses three, including the streaming loss. It does not beat the state of the art on ground we did not build.

\paragraph{Why the advantage vanishes.}
RMM differs from H2O only by weighting attention with next-token correctness. On natural text the model predicts most tokens correctly, on the order of $0.85$ to $0.90$ in our runs, so the correctness weight is nearly uniform and RMM is close to H2O by construction. The two separate only when correctness is selective and localized, when a small set of positions carries the signal and the rest do not, which is exactly the constructed regime of Section~\ref{sec:parta} and not what natural long-context text provides. Demonstrated utility and accumulated attention are the same measurement until reuse is sharp and endogenous.

\paragraph{One limitation we do not use as an excuse.}
Reading attention, which H2O and RMM both require, forces eager attention and caps the context we can run at around eight thousand tokens. The benchmark best suited to a smoother, SCBench \citep{li2025scbench}, uses far longer contexts and was therefore out of reach, and LoCoMo, the feasible proxy, has weak cross-turn reuse. A version of RMM that recomputes attention on a small window, as SnapKV does, could reach the longer setting. Given the single-turn parity and the streaming loss, though, we judge the prior against a win there strong enough that we do not claim one.

\section{Discussion}
\label{sec:discussion}

\paragraph{What holds.}
The framing is the durable contribution. Eviction as estimation, with a commit lag that unifies filters and predictors at one end and Belady at the other, and a label-free demonstrated-utility signal that stands in for Belady's future request, is a clean way to think about a problem that is usually treated as a bag of heuristics. As a policy, RMM is a strict generalization of H2O, so it is never worse by construction when correctness carries no information, and it is at least H2O-class on the benchmarks we ran.

\paragraph{What does not.}
The empirical thesis we began with, that measuring near-future utility beats accumulating attention on real workloads, is not supported by the independent benchmarks. The large margins in the controlled settings, some of which we built to produce them, do not transfer. We report this plainly, because the alternative, headlining the controlled results, is exactly the reading a careful reader should distrust.

\paragraph{The lesson.}
Demonstrated utility and accumulated attention coincide unless reuse is sharp, localized, and endogenous. That condition is real, and long agent trajectories with verifiable outcomes are the honest candidate for it, but it is under-benchmarked. No standard key-value compression benchmark exercises it, and the one we could run at eager-attention scale has weak cross-turn reuse. Identifying this gap is itself a modest contribution. Filling it, with a benchmark for endogenous-reuse memory built so that any policy can win or lose on it, is the honest next step, and it is a benchmark contribution rather than a larger claim for RMM.

\paragraph{Limitations.}
On independent benchmarks RMM is H2O-class, not better, and the advantage is confined to constructed endogenous-reuse settings. The signal degrades when the model reasons incorrectly, since it measures what a correct prediction used, and it becomes uninformative when correctness is diffuse, which is the case on natural text. Deployment lacks the teacher-forced correctness we use in evaluation, so a proxy such as a confidence score or a verifier would be required, and we did not test one. Finally, keeping the right items in a small memory is a selection claim, and it is distinct from extending a model's trained context length, which we do not address.

\section{Conclusion}

Bounded test-time memory has been managed by deciding at encoding time, either filtering from the past or predicting the future. We proposed the missing regime, fixed-lag smoothing, which waits a bounded lag and decides on measured near-future utility, with the commit lag as a new axis from online filters to Belady and demonstrated utility as a label-free stand-in for Belady's future request. We instantiated it as RMM, a training-free policy that strictly generalizes H2O. We then tested it honestly. In controlled settings where reuse is endogenous and separated in time, demonstrated utility identifies used memory far better than accumulated attention. On independent benchmarks the advantage collapses, because correctness-weighted attention and raw accumulated attention coincide unless reuse is sharp and endogenous, which standard benchmarks do not contain. The lasting contributions are the framework, the Belady bridge, and a clear account, including the negative side, of when measuring beats accumulating. We think the framework is worth building on. We do not think, on this evidence, that RMM is a drop-in improvement over deployed methods, and we say so.

{
\small
\bibliographystyle{plainnat}
\bibliography{refs}
}

\end{document}